# Which Sections of a Research Paper Best Reveal Its Research Methods? Evidence from Library and Information Science

**Fang, Qiuyu** Department of Information Management, Nanjing University of Science and Technology, China | fangqiuyu@njust.edu.cn

**Hao, Jiayi** Department of Information Management, Nanjing University of Science and Technology, China | haojiayi@njust.edu.cn

**Zhang, Chengzhi*** Department of Information Management, Nanjing University of Science and Technology, China | zhangcz@njust.edu.cn | *corresponding author

## ABSTRACT

Research methods are essential carriers of knowledge contribution in academic papers. Automatic multi-label classification of research methods can support knowledge services such as method retrieval, review generation, and research intelligence analysis. While existing studies primarily rely on titles and abstracts, abstracts often provide only limited methodological information, whereas utilizing full-text content faces challenges related to excessive length and information redundancy. Therefore, this paper proposes a segment combination strategy by partitioning the full-text content according to its physical postion. Using an annotated corpus of 1,954 full-text articles from three representative journals in Library and Information Science (*JASIST*, *LISR*, and *JDoc*), we evaluate the classification performance of various segments and their combinations across multiple models. Experimental results indicate that methodological information is distributed unevenly within the full-text content, with the middle-to-late and final segments exhibiting greater discriminative power. Furthermore, integrating bibliographic metadata with cross-segment combination strategies effectively enhances classification performance.

## INTRODUCTION

Research methods serve as a pivotal basis for understanding disciplinary development, evaluating academic rigor, and promoting paradigm evolution, representing an integral component of any scholarly field (Park, 2004). By systematically mining of research methods across various disciplines, researchers can reveal methodological preferences, evolutionary paths, and shifts in research focus over time. Such method-oriented analysis holds irreplaceable value for exploring research trends, refining knowledge systems, and enhancing the scientific nature of academic evaluation (Matusiak et al., 2025), thereby facilitating the progress of disciplines or fields.

Academic papers are the primary carriers of research achievements (Law et al., 1988). Compared with general text, academic literature exhibits distinct genre conventions and logical structure (Marius, 1991). Direct evidence for identifying research methods can be extracted from descriptions of research design, data sources, and analytical procedures. Consequently, academic papers are a crucial basis for analyzing methodological applications and innovations. In this context, the automated identification and classification of research methods have become a vital foundation for studies on knowledge graphs and disciplinary evolution. However, their practical implementaiton faces the dual challenges of the rapid growth of the literature and the increasing length of full-text content. Traditional manual coding is time-consuming and limited in sample size, making it difficult to support large-scale analysis (Grimmer & Stewart, 2013; Schmidt et al., 2024). While dictionary- or rule-based methods are convenient, they struggle to capture deeper contextual semantic relationships and require extensive manual rule design to achieve adequate coverage.

Recently, deep learning and Large Language Models (LLMs) have made significant progress in text understanding, providing new technical avenues for automated research method classification. Most existing studies rely on titles and abstracts; however, the methodological information contained in abstracts is often limited and lacks detailed descriptions of research design and operational procedures, which constrains classification accuracy for complex methods. Consequently, some scholars have begun to incorporate full-text content. While some attempts focus on identifying specific rhetorical sections (e.g., "Methodology" or "Conclusion"), the diversity of writing norms across disciplines and journals makes precise section identification difficult and limits the generalizability. Conversely, directly utilizing the entire full-text content for classification not only introduces substantial noise and increases computational costs but is also restricted by model context-window limitations, leading to the truncation of critical information. Therefore, efficiently selecting and combining text segments with the highest information gain from redundant full-text content has become a key step toward improving the stability and accuracy of automated research method classification.

Building on this, this study uses 1,954 full-text articles from three representative Library and Information Science (LIS) journals—*JASIST*, *LISR*, and *JDoc*—as the research sample, and proposes a segment combination strategy based on the physical position of the text within articles. Unlike logic-based segmentation, the proposed strategy does not attempt to identify rhetorical sections. Instead, it provides a simple, unified and reproducible way to organize full-text inputs. Based on this interval-based partitioning approach, mainstream models are fine-tuned using multiple segment combinations to investigate how different textual segments and their combinations affect the multi-label classification of research methods. The code and dataset for this paper can be accessed at *https://github.com/iiiiivan1/research-method-classification*.

## LITERATURE REVIEW

This chapter reviews the classification systems of research methods, automated classification techniques, and the utilization of structural information in LIS. It summarizes current research developments from both domestic and international perspectives.

### Overview of Research Methodology Frameworks in LIS

Systematic methodological research is essential for understanding disciplinary evolution, with robust classification frameworks serving as its foundation. In LIS, various scholars have developed distinct systems from diverse perspectives. Seminal work by Järvelin and Vakkari established a multi-dimensional framework covering research strategies and data collection, laying the cornerstone for methodological analysis(Jarvelin & Vakkari, 1990). Building on this, Chu developed a more granular taxonomy specifically for data collection, further refining the study of empirical processes(Chu, 2015; Chu & Ke, 2017).

Subsequent research has expanded these classic frameworks through empirical validation and optimization. Studies have confirmed the generalizability of existing systems across shifting paradigms and diverse corpora (Gauchi Risso, 2016; J. Ma & Lund, 2021). To accommodate emerging technologies, Zhang refined Chu's "experiment" category by incorporating computational methods such as classification and clustering(Zhang et al., 2021). Additionally, Jamali addressed boundary ambiguity in qualitative research by clarifying three often-confused methods, providing practical guidelines for manual annotation(Jamali, 2018). Collectively, while specific standards vary, these maturing frameworks share a consistent core logic and provide the essential evaluative criteria for the automated classification conducted in this study.

### Automated Research Method Classification

The classification of research methods has transitioned from manual coding to automated approaches. Historically, LIS methodological studies relied on manual content analysis, where researchers coded papers based on titles, abstracts, and full-text content(Ibrahim, 2021; Lou et al., 2021; Matusiak et al., 2025). While providing deep semantic understanding, manual methods are labor-intensive and unmeasurable for large-scale datasets. Early automation employed rule-based systems, dictionaries, or supervised learning (e.g., SVM, Random Forest) (Eckle-Kohler et al., 2013; Wilczynski et al., 2004; Zhang et al., 2021). However, these approaches often yield limited accuracy due to their reliance on local statistical features and inability to capture complex academic semantics(Eykens et al., 2021; Serenko, 2021; Wahid et al., 2021).

Recently, deep learning models, notably BERT, have enhanced classification performance through superior semantic representation (Devlin et al., 2019; Minaee et al., 2022; Zhang et al., 2023), yet they remain constrained by input length limits. Currently, the emergence of Large Language Models (LLMs), such as ChatGPT and DeepSeek, offers transformative potential. Leveraging advanced context awareness and reasoning capabilities, these models provide new avenues for extracting structured methodological insights from academic literature(Dagdelen et al., 2024; Foppiano et al., 2024)

### Research on the Application of Document Structural Information

Directly inputting full-text content introduces noise, and model performance often decays non-uniformly as input length increases(Idahl & Ahmadi, 2025; Zhu et al., 2025; Hong et al., 2025). Conversely, simple truncation leads to the loss of critical information(Radensky et al., 2026). Unlike general long-form text, the standardized nature of academic papers makes representing full-text semantics through the extraction of key segments feasible. Current research on document structure primarily bifurcates into logical and physical structural combinations.

Logical structure research typically utilizes the standard IMRaD framework (Burrough-Boenisch, 1999) to identify and combine specific rhetorical sections through natural language processing techniques. Subsequent studies have explored the varying contributions of these sections to specific tasks(Ma et al., 2022; Tan et al., 2025; Zhang et al., 2025). For instance, in automated novelty prediction, comparing 15 rhetorical section combinations revealed that local combinations outperformed full-text input (Wu et al., 2025). Others have improved abstract generation by identifying structural functions and capturing inter-sectional context (Bao et al., 2025). However, despite commonalities in knowledge systems, differences in organizational norms across disciplines remain, posing challenges for the precise and automated identification of logical structures.

In contrast, physical structure is intuitive and offers greater universality. Rather than relying on specific semantic labels, physical structure research analyzes text based on linear positions. By mapping full-text content to a linear space, studies have found that key background logic, research intent, and methodological details cluster within specific physical zones (e.g., the initial or final 10–15% of the text), a pattern that remains robust across disciplines and corpus scales (Bertin et al., 2016; Boyack et al., 2018). This approach obviates the need for complex section parsing. Consequently, this study designs and compares various physical segment combinations to identify intervals with the highest density of methodological features, aiming for a more universal and efficient approach to method classification.

## METHODOLOGY

Focusing on the LIS field, this study uses the full texts of academic papers to investigate methodological classification. Specifically, we first construct a research method classification corpus and perform data cleaning and preprocessing on the full-text papers. On this basis, we propose a series of segment combination schemes based on linear positional partitioning to exploit the physical structure and sequential organization of scholarly papers. By conducting supervised fine-tuning on LLMs and pre-trained encoders, we develop multi-label classification models for research methods and systematically evaluate the informational utility of distinct positional segments. The comprehensive research workflow is depicted in Figure 1.

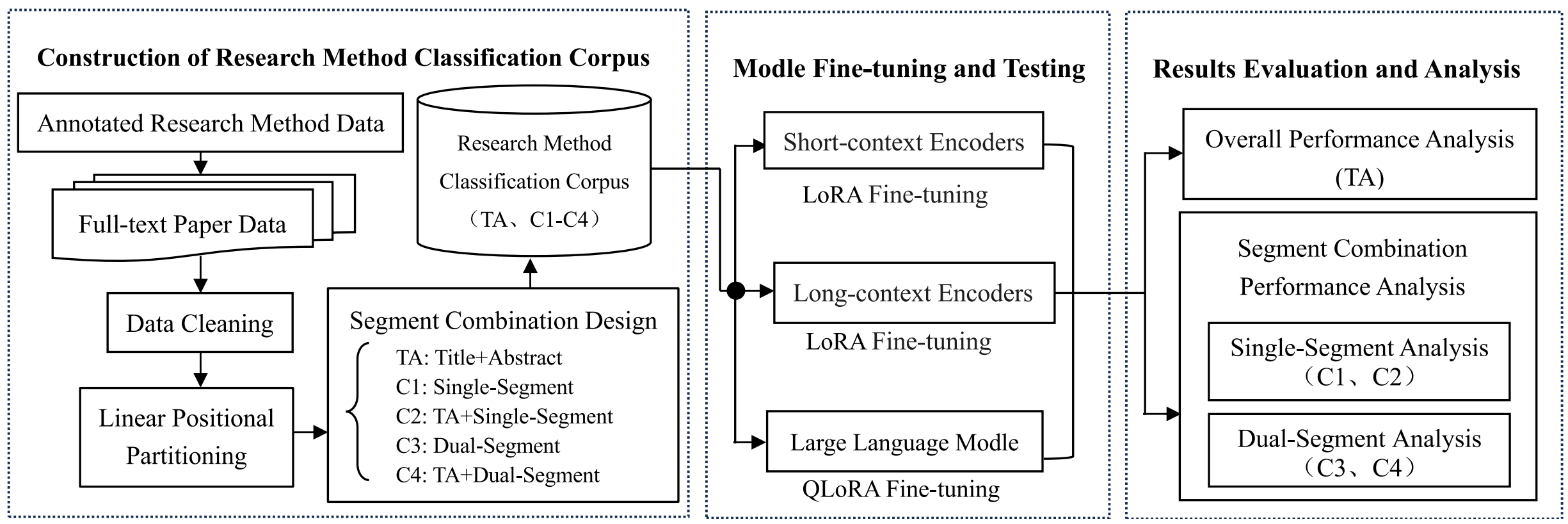

***Note:*** A single-segment refers to an individual body text interval derived from the linear positional partitioning of the paper; a dual-segment represents a combination of two segments from distinct positions.

**Figure 1. Framework of This Study**

### Dataset Construction and Preprocessing

This study utilizes the annotated data from Chu and Ke as its primary source(Chu & Ke, 2017). The corpus consists of 1,981 research articles published between 2001 and 2010 in three journals: *Journal of Documentation* (*JDoc*), *Journal of the Association for Information Science and Technology* (*JASIST*, formerly the *Journal of the American Society for Information Science and Technology*), and *Library & Information Science Research* (*LISR*). The original annotation process reported an intercoder agreement rate of 86.7%, exceeding the commonly accepted threshold of 80%, indicating the reliability of the method labels used in this study. The methodological taxonomy employed is presented in Table 1. Primarily categorized by data collection methods, this system comprises 16 method types and covers the mainstream research methodologies in the field of LIS.

Full-text data in HTML format were collected based on the annotated records. Through format conversion and preprocessing, raw data were extracted into plain text (TXT). We leveraged the hierarchical structure of HTML (e.g., heading tags and paragraph boundaries) to perform structural parsing and field annotation for elements such as titles, abstracts, and level-1 headings. As shown in Table 2, the number of level-1 headings varies across articles, indicating differences in article organization and structural granularity. Research method labels were mapped to a unified <method> field, and noisy segments—including acknowledgments, references, and appendices—were removed. To resolve issues such as garbled characters and formatting errors post-conversion, we conducted manual auditing and cleaning. Furthermore, to ensure the validity of subsequent experiments based on discourse structure, non-standard articles (e.g., those missing abstracts or level-1 headings) were excluded. Ultimately, a dataset comprising 1,954 academic papers for research method classification was constructed.

| No. | Method Name | No. | Method Name |
|---|---|---|---|
| $M_1$ | Bibliometrics | $M_9$ | Observation |
| $M_2$ | Content analysis | $M_{10}$ | Questionnaire |
| $M_3$ | Delphi study | $M_{11}$ | Research diary/journal |
| $M_4$ | Ethnography/field study | $M_{12}$ | Theoretical approach |
| $M_5$ | Experiment | $M_{13}$ | Think-aloud protocol |
| $M_6$ | Focus groups | $M_{14}$ | Transaction log analysis |
| $M_7$ | Historical method | $M_{15}$ | Webometrics |
| $M_8$ | Interview | $M_{16}$ | Other methods |

**Table 1. Research Method Taxonomy**

| Number of Level-1 Headings | Number of Articles | Percentage | Number of Level-1 Headings | Number of Articles | Percentage |
|---|---|---|---|---|---|
| 2-3 | 42 | 2.15% | 7-9 | 759 | 38.84% |
| 4-6 | 1004 | 51.38% | 10-25 | 149 | 7.63% |

**Table 2. Distribution of Level-1 Headings**

## Construction of Experimental Corpora for LIS Research Method Classification

### *Linear Position Partitioning and Combination Schemes*

To investigate the distribution of methodological information throughout the full-text content and its contribution to classification, this study constructs multiple sets of comparable experimental inputs from the preprocessed corpus. As shown in Table 2, the number of level-1 headings varies substantially across articles. To ensure that this structural heterogeneity was not merely driven by one specific journal, we further examined the journal-level distribution of heading counts. The results indicate that variation in article organization exists across all three selected journals. Therefore, directly segmenting papers by level-1 headings would make it difficult to construct uniform and comparable inputs. In addition, paragraph-based segmentation faces a similar problem because natural paragraphs are fine-grained units whose number and length may vary considerably across papers. For this reason, this study adopts a linear physical-position partitioning strategy(in Figure 2), which divides each article according to proportional text positions rather than variable structural units.To ensure comparability across experimental groups, the classification results of the "Title + Abstract" combination serve as the baseline, hereafter referred to as TA.

Inspired by Bertin, this study adopts a character-based equal-division method from a linear perspective(Bertin et al., 2016). Specifically, the body text is first normalized into a continuous character sequence and divided into ten theoretical cut-points based on the total character length. Local detection windows are then established around each cut-point to fine-tune the boundaries using white spaces and end-of-sentence punctuation, thereby avoiding word or sentence fragmentation. This process divides the body text into ten segments, each representing approximately 10% of the content, denoted as $\{S_1, S_2, \ldots, S_{10}\}$. Subsequently, four sets of combination schemes (C1–C4) were designed:

**(1) Single-Segment Input (C1):** Each individual segment $S_i$ is used independently as model input for fine-tuning and evaluation.

**(2) TA-Augmented Segment Input (C2):** Bibliographic metadata (TA) is concatenated with a single segment to form ten experimental inputs, denoted as TA+$S_i$.

**(3) Dual-Segment Input (C3):** Pairs of segments from different positions are concatenated(e.g. $S_i \oplus S_j$) to evaluate their combined effect.

**(4) TA-Augmented Dual-Segment Input (C4):** Bibliographic metadata (TA) is combined with two different positional segments, forming the input TA+$S_i \oplus S_j$.

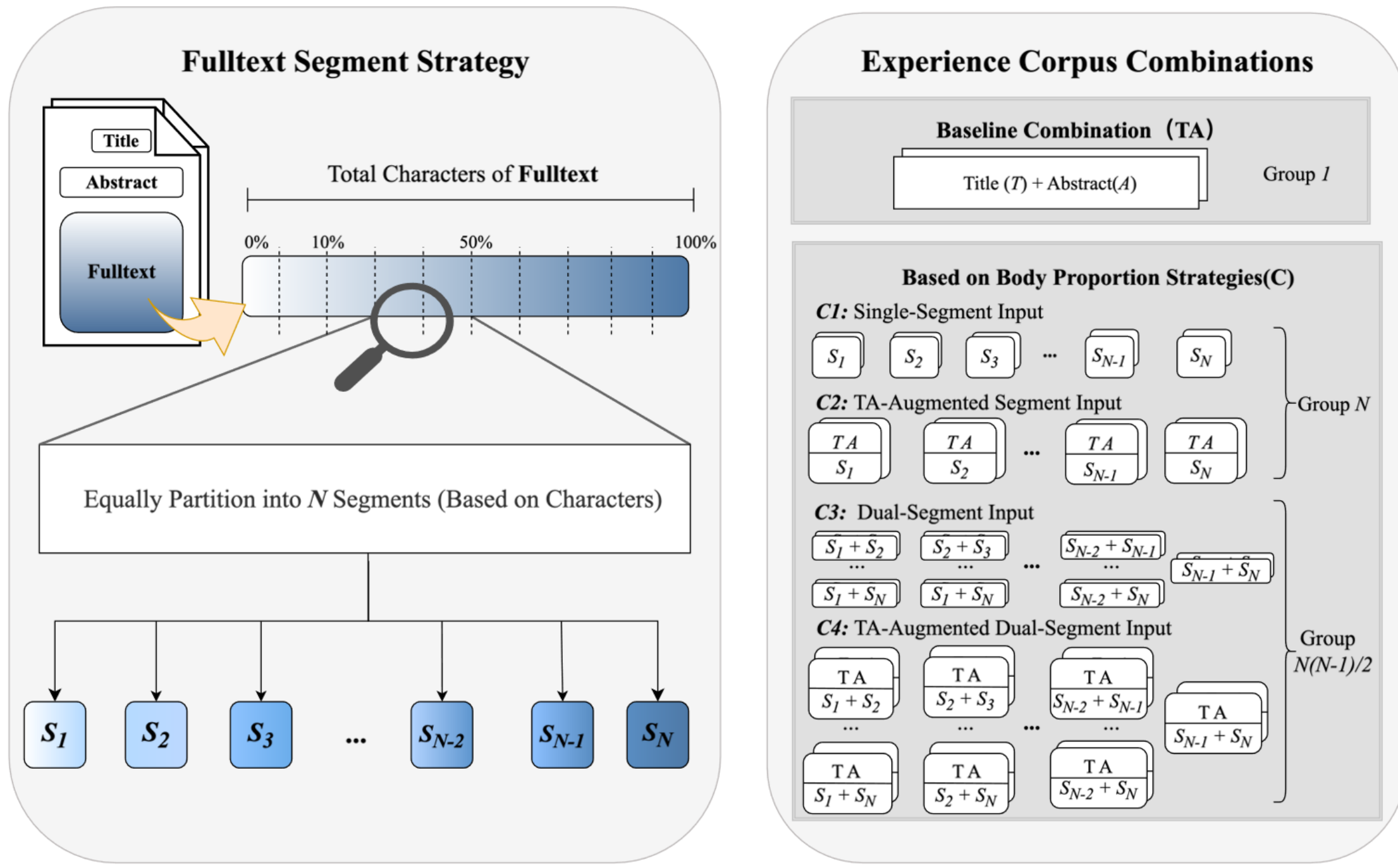

***Note*:** In this paper, $N$=10.

**Figure 2. Schematic Diagram of Segment Combination Schemes Based on Physical Structure**

### *Methodological Summarization for Long Text Segments*

Considering the 512-token input constraint of BERT and SciBERT models(Beltagy et al., 2019; Devlin et al., 2019), most segment combinations constructed in this study significantly exceed this limit. To ensure informational integrity across different models, we introduce "methodological summaries" as compressed input content. A "method summary" refers to a concise text segment derived by extracting and distilling research-method-related information from the original text, specifically preserving core methodological elements such as research design, data sources, sampling information, analytical processes, tools/models, and evaluation metrics.

In practice, DeepSeek-V3.2 was employed for semantic compression and summarization, with the temperature parameter set to 0. As illustrated in Figure 3, we designed a structured prompt to guide the model in extracting key methodological evidence and reconstructing it into a summary of approximately 220 words. Manual evaluation confirmed that the summaries generated by DeepSeek are highly reliable and effectively filter out redundant information from the corpus.

**PROMPT**

From Title and Text (Abstract+Segment), internally do 2 steps:
(1) Extract and compress ONLY explicitly stated methodological evidence (design/procedure, data/samples, tools/software, preprocessing, models/algorithms, parameters/variables, evaluation metrics) into ≤400 words.(2) Using ONLY that evidence, write a method-focused abstract of about 220 words (target range 180–250 words), prioritizing methodological details (data source, sample scope, coding/classification procedure, variables, and analysis steps).
Rules: no judgment/classification/inference, do NOT mention missing/unknown info, keep variable names/parameters EXACTLY as written, do NOT invent symbols/notation, prefer concrete methodological details over background/context。If the text contains little explicit methodological evidence, write a concise neutral summary of the segment content instead of returning None.
Output EXACTLY 2 lines:title: <exact Title>, abs: <method-focused abstract OR concise neutral summary>

**OUTPUT**

title: Subject knowledge improves interactive query expansion assisted by a thesaurus
abs: This study recorded the search logs and conducted pre- and post-search interviews of 15 expert undergraduates in pedagogy and 15 novice students with no previous studies in the field…

**Figure 3. Illustration of the Prompt for Methodological Summary Generation**

## Construction of Research Method Classification Models

This study formulates research method classification as a multi-label text classification task. For a given segment combination input *X*, the goal is to predict the corresponding set of methodological labels $Y \subseteq \{M_1, M_2, ..., M_{16}\}$. To examine the applicability of our positional segment strategy across various architectures, we constructed three categories of models: short-context encoders, long-context encoders, and Large Language Models, and designed corresponding input processing strategies to address the window constraints specific to each architecture.

For short-context models with a 512-token constraint (BERT and SciBERT)—where BERT is a bidirectional Transformer encoder and SciBERT is pre-trained on scientific literature to better suit scholarly text scenarios(Beltagy et al., 2019; Devlin et al., 2019), a "fair truncationstrategy" was employed for cross-segment combinations. This strategy equally allocates the token budget among all segments to minimize information imbalance resulting from truncation. For long-context models with a 4096-token window (SciBERT-long and BigBird), distinct implementations were adopted to accommodate their different long-sequence modeling mechanisms while maintaining consistent input organization principles. Specifically, for SciBERT-long, global attention masks were constructed to enhance the model's global aggregation capability; for BigBird, its sparse attention mechanism was leveraged to model long-range dependencies (Beltagy et al., 2020; Zaheer et al., 2020). For LLMs (Llama-3-8B, DeepSeek-LLM-7B, and Qwen-7B), we adopted structured concatenation with delimiters and input alignment to ensure a balanced representation of information from different positions.

Regarding the encoder models, a linear classification head was integrated atop the final [CLS] representation, using a Sigmoid activation function to achieve independent probability predictions for multi-label output. During training, Parameter-Efficient Fine-Tuning (PEFT) via LoRA was implemented to reduce the scale of trainable parameters (Hu et al., 2022). For LLMs, we utilized a controlled generation approach. Titles, abstracts, and segments were organized into structured inputs via delimiters, guiding the model to generate a list of method labels, which were then mapped to a predefined label set to ensure consistency with the encoder-based results.

# EXPERIMENTS AND RESULTS ANALYSIS

This chapter presents the analysis of results for the automated research method classification task based on textual segment combinations. First, the experimental setup and evaluation metrics are introduced, followed by a comparative analysis of the classification performance across different segment combination schemes. Finally, the underlying influence mechanisms are discussed through specific case studies.

## Experimental Settings

### *Data Processing*

The corpus of 1,954 papers was randomly partitioned into training, validation, and test sets using an 8:1:1 ratio. A fixed random seed was used throughout all experiments to ensure reproducibility. To ensure comparability across experimental groups C1–C4, identical random split indices were utilized, guaranteeing that each paper was assigned to the same subset in every experimental group. Based on these indices, four sets of combined training, validation, and test samples were generated for independent training and evaluation. For length-constrained models, we compared two input formats: "direct truncation" and "methodological summaries." For long-context models and LLMs, raw inputs and structured template-based inputs were employed, respectively. All prediction results were mapped to a predefined label space to establish a unified performance evaluation standard.

### *Parameter Settings and Evaluation Metrics*

During the fine-tuning phase, encoder-based models utilized Parameter-Efficient Fine-Tuning (PEFT) via LoRA. For short-context encoders, the maximum input length was set to 512, with r=64 and a scaling factor $\alpha = 128$; a fair length allocation strategy was applied to segment combinations requiring truncation. For long-context encoders, the maximum input length was set to 4,096, with r=32 and $\alpha = 64$. Specifically, SciBERT-long incorporated a global attention mask, while BigBird utilized a sparse attention mechanism. For LLMs, we selected Llama-3-8B-Instruct, DeepSeek-LLM-7B-Chat, and Qwen2.5-7B-Instruct for supervised fine-tuning (SFT) using QLoRA. These models were loaded with 4-bit quantization and bfloat16 computational precision, with a maximum length of 4,096, a LoRA rank of r=16, and a scaling factor $\alpha = 32$.

Given the nature of multi-label research method classification, we adopted standard metrics: *Micro-averaged Precision* (*Micro-P)*, *Micro-averaged Recall* (*Micro-R*), and *Micro-averaged $F_1$-score* (*Micro-$F_1$*). The calculation formulas are as follows:

$$Micro\text{-}P = \frac{\sum_{k=1}^{M} TP_k}{\sum_{k=1}^{M} (TP_k + FP_k)} \tag{1}$$

$$Micro\text{-}R = \frac{\sum_{k=1}^{M} TP_k}{\sum_{k=1}^{M} (TP_k + FN_k)} \quad (2)$$

$$Micro\text{-}F_1 = \frac{2 \times Micro\text{-}P \times Micro\text{-}R}{Micro\text{-}P + Micro\text{-}R} \quad (3)$$

Where $M$ denotes the number of research method categories; $TP_k$ and $FP_k$ represent the true positives and false positives for the *k-th* category, respectively; and $FN_k$ represents the false negatives for the *k-th* category. To balance accuracy and coverage, the $Micro\text{-}F_1$ score serves as the primary metric for evaluating classification performance.

## Analysis of Classification Results Under Different Positional Segment Combinations

### *Performance Comparison Under the Baseline Combination*

In this section, we conduct a horizontal comparison of the multi-label classification performance across all models using the baseline combination (TA: Title+Abstract). The classification results are presented in Table 3.

| Metrics | BERT | SciBERT | SciBERT-long | BigBird | DeepSeek-LLM-7B | Llama-3-8B | Qwen2.5-7B |
|---|---|---|---|---|---|---|---|
| *Micro_P* | 0.6679 | 0.6981 | **0.8308** | 0.7402 | 0.6680 | 0.7041 | 0.6078 |
| *Micro_R* | 0.6434 | **0.7088** | 0.6278 | 0.5875 | 0.6420 | 0.5370 | 0.6031 |
| $Micro_F_1$ | 0.6554 | 0.7034 | **0.7152** | 0.6551 | 0.6548 | 0.6093 | 0.6055 |

**Table 3. Performance of Models on the Baseline TA Combination**

Overall, encoder-based pre-trained models outperformed LLMs. Specifically, SciBERT-long achieved the best overall performance with an $F_1$-score of 0.7152, followed by SciBERT at 0.7034. These results suggest that when using only titles and abstracts as input, both domain-specific pre-training and long-context modeling are more effective at capturing features relevant to research methods. BigBird and BERT yielded moderate results, with *F1-scores* of 0.6551 and 0.6554, respectively, indicating certain limitations of general-purpose pre-trained models in identifying methodological characteristics.Among the three LLMs, DeepSeek-LLM-7B performed the best but still fell short of SciBERT-long. Notably, Llama-3-8B exhibited higher precision but lower recall, reflecting a tendency toward conservative output. These baseline results establish a unified reference for the subsequent gain analysis following the introduction of various body-text positional segments.

### *Comparative Analysis of Single Segments and TA-Augmented Combinations*

Following the performance comparison under the baseline combination, this section further explores the contribution of specific positional segments to classification. Figure 4 illustrates the performance of single-segment input (C1) and TA-augmented segment input (C2) across various segments.

As shown in Figure 4(a), the $F_1$-scores for C1 fluctuate across different positional segments, confirming that methodological features are non-uniformly distributed within the physical structure of academic papers. Overall, most C1 inputs perform below the baseline, indicating that titles and abstracts possess a higher information density of research method characteristics than any single body-text segment. Notably, most models exhibit a performance trough near i=2 and peaks near i=6 and i=10. This suggests that the middle-to-late and concluding intervals of the body text are more likely to contain essential methodological elements, such as research design, data sampling, experimental settings, and evaluation metrics.

Figure 4(b) reveals that the C2 scheme (incorporating bibliographic metadata) generally improves performance over C1 across most segments. This demonstrates that when using a single body-text segment, the inclusion of TA effectively enhances classification results. However, compared to the baseline combination, the overall performance of C2 shows no significant improvement; in some segments, performance even stagnated or slightly declined due to local noise interference. This phenomenon suggests that methodological clues are dispersed across multiple text intervals, and a single positional segment is insufficient to capture the critical information required for comprehensive classification. Consequently, to better utilize features across different locations, the next section will analyze the performance of cross-positional dual-segment combinations.

Furthermore, for short-context encoders, the introduction of methodological summaries in C1 and C2 did not yield stable performance gains. This implies that the summarization process may blur certain fine-grained feature words, thereby negatively impacting the model’s discriminative capability.

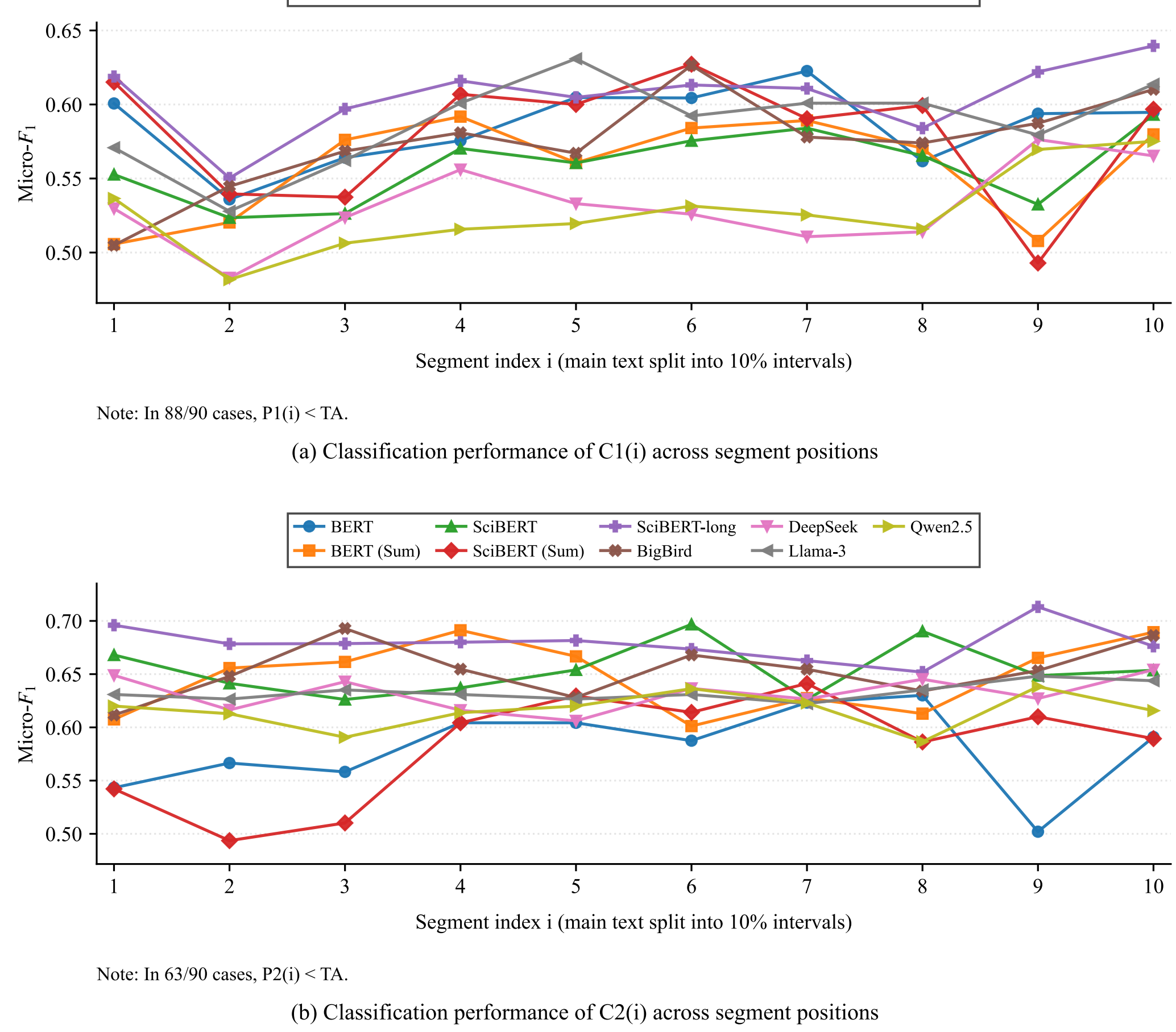

(a) Classification performance of C1(i) across segment positions

(b) Classification performance of C2(i) across segment positions

**Figure 4. Comparison of classification performance across different positional segments under C1 and C2 inputs**

*Comparative Analysis of Dual-Segment Combinations and TA-Augmented Dual-Segment Combinations*
To more comprehensively utilize the methodological features within text segments, this section further explores the classification performance of cross-positional dual-segment combinations (C3) and their TA-augmented dual-segment combinations (C4).

As shown in Table 4, compared to the single-segment input (C1), the optimal C3 combinations demonstrate higher classification performance across all models, with several models even surpassing the TA baseline. This indicates that combining two segments from different positions enables the model to acquire more comprehensive methodological features. Further observation of the distribution of optimal combinations reveals that high-performing pairs often originate from the "middle-to-late+final" or "middle+middle-to-late" intervals. This suggests that these specific intervals likely carry complementary methodological information.The addition of bibliographic metadata (TA) to these dual-segment combinations (C4) resulted in further performance gains for most models, with the majority of optimal C4 combinations exceeding the TA baseline. This result demonstrates that, compared to the limited information provided by single-segment combinations, the fusion of cross-positional dual segments with TA provides not only essential background context but also detailed evidence regarding research design and data processing, thereby enhancing the model's discriminative power.

Additionally, the use of methodological summaries for input segments exhibited strong compatibility within the C3 and C4 schemes. Summarized versions were more likely to achieve higher optimal performance than their raw text counterparts. This suggests that when dealing with longer and more redundant information, methodological summarization helps highlight key methodological elements while mitigating noise interference. Notably, for SciBERT, the summarized version of the TA+ $S_5 + S_6$ combination achieved the highest performance across all experimental groups. This underscores that for domain-specific pre-trained models, the integration of distilled local details with bibliographic metadata yields the best results under the current data and experimental settings.

| Model | TA_$F_1$ | Best_C3 | | | | Best_C4 | | | |
|---|---|---|---|---|---|---|---|---|---|
| | | pair | $F_1$ | *P* | *R* | pair | $F_1$ | *P* | *R* |
| BERT | 0.6554 | 7-10 | 0.6419 | 0.6479 | 0.6360 | 4-9 | 0.6389 | 0.6940 | 0.5919 |
| BERT（Sum） | 0.6554 | 5-8 | 0.6529 | 0.7453 | 0.5809 | 7-8 | 0.6973 | 0.8068 | 0.6140 |
| SciBERT | 0.7034 | 5-10 | 0.6439 | 0.6641 | 0.6250 | 6-7 | 0.7166 | 0.7973 | 0.6507 |
| SciBERT（Sum） | 0.7034 | 6-10 | 0.6639 | 0.7559 | 0.5919 | **5-6** | **0.7208** | **0.7403** | **0.7022** |
| SciBERT-long | 0.7152 | 4-10 | 0.6861 | 0.7895 | 0.6066 | 4-9 | 0.7057 | 0.7018 | 0.7096 |
| Bigbird | 0.6551 | 7-10 | 0.6541 | 0.6732 | 0.6360 | 6-10 | 0.6341 | 0.6778 | 0.5956 |
| DeepSeek-LLM-7B | 0.6548 | 3-10 | 0.6472 | 0.7416 | 0.5741 | 5-10 | 0.7107 | 0.8037 | 0.6370 |
| Llama-3-8B | 0.6093 | 2-10 | 0.6652 | 0.7908 | 0.5741 | 1-4 | 0.6652 | 0.7908 | 0.5741 |
| Qwen2.5-7B | 0.6055 | 2-10 | 0.6269 | 0.7387 | 0.5444 | 6-10 | 0.6806 | 0.7799 | 0.6037 |

**Table 4. Optimal Performance and Corresponding Segment Combinations Under C3 and C4 Schemes**

To assess whether the performance improvements achieved by the segment combination strategies were statistically robust, we conducted an approximate randomization test, implemented as a Monte Carlo paired permutation test(Dror et al., 2018; Noreen, 1989), with 5,000 random swaps on the fixed test set. Statistical significance was evaluated based on the differences in *Micro-*$F_1$ scores between the optimal C4 schemes and the Title+Abstract (TA) baseline. Since the detailed *Micro-*$F_1$ values have been reported in Table 4, Table 5 reports only the *Micro-*$F_1$ differences and the corresponding p-values. The results show that the optimal C4 schemes significantly outperform the TA baseline for BERT, DeepSeek-LLM-7B, Llama-3-8B, and Qwen2.5-7B. For SciBERT, although C4 achieves a positive *Micro-*$F_1$ gain, the improvement does not reach statistical significance.

| Metrics | BERT | SciBERT | DeepSeek-LLM-7B | Llama-3-8B | Qwen2.5-7B |
|---|---|---|---|---|---|
| ***Comparison*** | TA vsC4(7-8) | TA vs C4(5-6) | TA vs C4(5-10) | TA vs C4(1-3) | TA vs C4(6-10) |
| ***ΔMicro-*$F_1$** | 0.0393 | 0.0174 | 0.0559 | 0.0559 | 0.0751 |
| ***p-value*** | 0.0326 | 0.3416 | 0.0118 | 0.0002 | 0.0120 |
| ***Sig.*** | * | n.s. | * | * | * |

**Table 5. Statistical significance test results for improvements over the TA baseline**

***Note.*** * indicates $p < 0.05$; n.s. indicates not significant. The approximate randomization test was conducted with 5,000 random swaps on the fixed test set.

## Case Study

To further reveal how different positional segment combinations influence classification performance, this section selects a study on communication patterns in academic library online consultations as a case study. The ground truth method label for this article is "Content Analysis."

As illustrated in Figure 5, SciBERT misclassified the article as "Questionnaire" when provided only with the title and abstract (TA). Analysis of the input reveals that although the abstract mentions the term "Discourse Analysis," it lacks specific details regarding coding items, coding procedures, and analytical steps. Furthermore, the model was likely misled by quantitative expressions such as "402 chat transcripts (covering 6,572 minutes)" and "indicators" causing it to favor a common questionnaire-based data collection and analysis workflow. This suggests that models struggle to make stable and accurate classifications when faced with ambiguous terminology and interfering quantitative features in the abstract.

In contrast, under the TA-augmented dual-segment combination (utilizing methodological summaries), the model successfully corrected this misclassification. By incorporating segments S5 and S6 from the middle of the body text, specific coding details—such as "The procedure involved coding syntactic formality markers" and "content-based markers"—were introduced. these operational details clearly point to the procedural aspects of content analysis, effectively compensating for the macro-level descriptions in the abstract.This case demonstrates that while titles and abstracts tend to focus on macro-level descriptions of research questions and objects, the middle segments of the body text supplement critical details at the methodological implementation level. This enables cross-positional information complementarity from "research background" to "operational specifics," thereby mitigating interference from generalized expressions in the abstract and providing more robust evidence for precise method identification.

Furthermore, this finding validates the effectiveness of the proposed physical segment combination strategy in capturing deep methodological information.

| **title:** Chat reference communication patterns and implications: applying politeness theory | | | |
|---|---|---|---|
| TA（Title+Abstract） | **Pred : Questionnaire** | C4( $TA + S_5 + S_6$ ) | **Pred : Content Analysis** |
| **abstract:** Purpose … Design/methodology/approach This discourse analysis identifies the syntactic and content indicators and patterns of formality levels in a purposive sample of 402 chat transcripts (covering 6,572 minutes) from one academic year at a large, US, public university. Findings Syntactic formality markers include regular use of contractions, slang, sentence fragments, and non-standard punctuation as well as limited use of acronyms and abbreviations with rare use of emoticons. Content-based markers included apologies, self-disclosure, and expressions of extreme need. Use patterns are related to the level of responsibility assumed by the librarian as well as the interview stages. Research limitations/implications … **method:** Content Analysis | | **abs:** This discourse analysis used a purposive sample of 402 chat transcripts (6,572 minutes) from one academic year at a large US public university. The procedure involved coding syntactic formality markers (contractions, slang, sentence fragments, non-standard punctuation, acronyms, abbreviations, emoticons) and content-based markers (apologies, self-disclosure, expressions of extreme need). Patterns of formality level use were identified in relation to librarian-assumed responsibility and interview stages. **method:** Content Analysis | |

**Figure 5. Comparison of Classification Results for the Case Study Article under TA and C4 Inputs**

## CONCLUSION AND FUTURE WORKS

This study addresses the task of automated multi-label classification of research methods in academic literature. From the perspective of physical document structure, we proposed a strategy involving linear position-based partitioning and segment combinations. Comparative evaluations across multiple models were conducted to reveal the contribution patterns of different body-text intervals and their combinations toward method identification.

The results indicate that methodological features are non-uniformly distributed across the physical structure of academic papers. Segments in the middle-to-late and concluding intervals—which are more likely to carry implementation details and concluding remarks—exhibit higher discriminative value. By constructing TA-augmented dual-segment combinations (C4), we integrated global context from titles and abstracts with local methodological details, achieving superior performance over the baseline across most models. The distribution of optimal combinations reveals that high-performing pairs tend to cluster in the "middle+middle-to-late" or "middle-to-late+final" intervals, suggesting that the methodological evidence across different sections is inherently complementary. Furthermore, the experiments demonstrate that LLM-based semantic compression effectively filters out redundant noise, providing a promising new pathway for long-document classification.

However, certain limitations remain. First, this study utilized articles from only three journals in the field of LIS. Given the variations in writing norms and argumentative structures across different disciplines, the generalizability of these findings requires further validation across a broader range of fields. Second, while our reliance on physical structure ensures universality and reproducibility, linear partitioning does not always align perfectly with logical sections, which may compromise semantic integrity at segment boundaries. Future research could incorporate logic-based segmentation based on section functions and further optimize methodological summarization mechanisms to provide more interpretable segment selection criteria for methodological identification in long documents.

## GENERATIVE AI USE

We employed a generative AI tool for the following purpose: translating the manuscript into English. We evaluated the output through careful manual review and verification to ensure the accuracy, clarity, and appropriateness of the translation. The authors assume all responsibility for the content of this submission.

## AUTHOR ATTRIBUTION

Fang Q: methodology, data curation, formal analysis, writing; Hao J: writing – review and editing; Zhang C: conceptualization, methodology, formal analysis, funding acquisition, supervision, writing – review and editing.

## ACKNOWLEDGMENTS

This work was supported by the National Natural Science Foundation of China (Grant No.72074113), Major Projects of National Social Science Fund (Grant No. 25&ZD298).

## REFERENCES

Bao, T., Zhang, H., & Zhang, C. (2025). Enhancing abstractive summarization of scientific papers using structure information. Expert Systems with Applications, 261, 125529.

Beltagy, I., Lo, K., & Cohan, A. (2019). SciBERT: A pretrained language model for scientific text. Proceedings of the 2019 Conference on Empirical Methods in Natural Language Processing and the 9th International Joint Conference on Natural Language Processing (EMNLP-IJCNLP), 3615–3620.

Beltagy, I., Peters, M. E., & Cohan, A. (2020). Longformer: The Long-Document Transformer (arXiv:2004.05150). arXiv.

Bertin, M., Atanassova, I., Gingras, Y., & Larivière, V. (2016). The invariant distribution of references in scientific articles. Journal of the Association for Information Science and Technology, 67(1), 164–177.

Boyack, K. W., van Eck, N. J., Colavizza, G., & Waltman, L. (2018). Characterizing in-text citations in scientific articles: A large-scale analysis. Journal of Informetrics, 12(1), 59–73.

Burrough-Boenisch, J. (1999). International reading strategies for IMRD articles. Written Communication, 16(3), 296–316.

Chu, H. (2015). Research methods in library and information science: A content analysis. Library & Information Science Research, 37(1), 36–41.

Chu, H., & Ke, Q. (2017). Research methods: What's in the name? Library & Information Science Research, 39(4), 284–294.

Dagdelen, J., Dunn, A., Lee, S., Walker, N., Rosen, A. S., Ceder, G., Persson, K. A., & Jain, A. (2024). Structured information extraction from scientific text with large language models. Nature Communications, 15(1), 1418.

Devlin, J., Chang, M.-W., Lee, K., & Toutanova, K. (2019). Bert: Pre-training of deep bidirectional transformers for language understanding. Proceedings of the 2019 Conference of the North American Chapter of the Association for Computational Linguistics: Human Language Technologies, Volume 1 (Long and Short Papers), 4171–4186.

Dror, R., Baumer, G., Shlomov, S., & Reichart, R. (2018). The hitchhiker's guide to testing statistical significance in natural language processing. Proceedings of the 56th Annual Meeting of the Association for Computational Linguistics (Volume 1: Long Papers), 1383–1392.

Eckle-Kohler, J., Nghiem, T., & Gurevych, I. (2013). Automatically assigning research methods to journal articles in the domain of social sciences. Proceedings of the American Society for Information Science and Technology, 50(1), 1–8.

Eykens, J., Guns, R., & Engels, T. C. (2021). Fine-grained classification of social science journal articles using textual data: A comparison of supervised machine learning approaches. Quantitative Science Studies, 2(1), 89–110.

Foppiano, L., Lambard, G., Amagasa, T., & Ishii, M. (2024). Mining experimental data from materials science literature with large language models: An evaluation study. Science and Technology of Advanced Materials: Methods, 4(1), 2356506.

Gauchi Risso, V. (2016). Research methods used in library and information science during the 1970-2010. New Library World, 117(1–2), 74–93.

Grimmer J., & Stewart B. M. (2013). Text as Data: The Promise and Pitfalls of Automatic Content Analysis Methods for Political Texts. Political Analysis, 21(3), 267–297.

Hu, E. J., Shen, Y., Wallis, P., Allen-Zhu, Z., Li, Y., Wang, S., Wang, L., & Chen, W. (2022). Lora: Low-rank adaptation of large language models. Iclr, 1(2), 3.

Ibrahim, B. (2021). Statistical methods used in Arabic journals of library and information science. Scientometrics, 126(5), 4383–4416.

Idahl, M., & Ahmadi, Z. (2025). Openreviewer: A specialized large language model for generating critical scientific paper reviews. Proceedings of the 2025 Conference of the Nations of the Americas Chapter of the Association for Computational Linguistics: Human Language Technologies (System Demonstrations), 550–562.

Jamali, H. R. (2018). Does research using qualitative methods (grounded theory, ethnography, and phenomenology) have more impact? Library & Information Science Research, 40(3–4), 201–207.

Jarvelin, K., & Vakkari, P. (1990). Content analysis of research articles in library and information science. Library and Information Science Research, 12(4), 395–421.

Law, J., Bauin, S., Courtial, J., & Whittaker, J. (1988). Policy and the mapping of scientific change: A co-word analysis of research into environmental acidification. Scientometrics, 14(3–4), 251–264.

Lou, W., Su, Z., He, J., & Li, K. (2021). A temporally dynamic examination of research method usage in the Chinese library and information science community. Information Processing & Management, 58(5), 102686.

Ma, B., Zhang, C., Wang, Y., & Deng, S. (2022). Enhancing identification of structure function of academic articles using contextual information. Scientometrics, 127(2), 885–925.

Ma, J., & Lund, B. (2021). The evolution and shift of research topics and methods in library and information science. Journal of the Association for Information Science and Technology, 72(8), 1059–1074.

Marius, R. (1991). Genre analysis: English in academic and research settings. JSTOR. https://www.jstor.org/stable/20865811

Matusiak, K. K., Osinska, V., Organisciak, P., & Thomas Pitts, R. (2025). Research methods and the use of visual representation in library and information science research. Journal of the Association for Information Science and Technology, 76(3), 527–544.

Minaee, S., Kalchbrenner, N., Cambria, E., Nikzad, N., Chenaghlu, M., & Gao, J. (2022). Deep Learning--based Text Classification: A Comprehensive Review. ACM Computing Surveys, 54(3), 1–40.

Noreen, E. W. (1989). Computer-intensive methods for testing hypotheses. Wiley New York. https://www.jeffreycjohnson.org/app/download/764734156/cimeth.pdf

Park, S. (2004). The study of research methods in LIS education: Issues in Korean and U.S. universities. Library & Information Science Research, 26(4), 501–510.

Radensky, M., Shahid, S., Fok, R., Siangliulue, P., Hope, T., & Weld, D. S. (2026). Human-LLM Compound System for Scientific Ideation through Facet Recombination and Novelty Evaluation (arXiv:2409.14634). arXiv.

Schmidt, J.-H., Goutier, M., Koch, L., Schwinghammer, R., & Benlian, A. (2024). Automatic Classification of IS Research Papers: A Design Science Approach. https://aisel.aisnet.org/ecis2024/track02_general/track02_general/8/

Serenko, A. (2021). A structured literature review of scientometric research of the knowledge management discipline: A 2021 update. Journal of Knowledge Management, 25(8), 1889–1925.

Tan, H., Zhan, S., Jia, F., Zheng, H.-T., & Chan, W. K. V. (2025). A hierarchical framework for measuring scientific paper innovation via large language models. Information Sciences, 122787.

Wahid, N., Warraich, N. F., & Tahira, M. (2021). Group level scientometric analysis of Pakistani authors. COLLNET Journal of Scientometrics and Information Management, 15(2), 287–304.

Wilczynski, N. L., Haynes, R. B., & the Hedges Team, (2004). Developing optimal search strategies for detecting clinically sound prognostic studies in MEDLINE: An analytic survey. BMC Medicine, 2(1), 23.

Wu, W., Zhang, C., Bao, T., & Zhao, Y. (2025). SC4ANM: Identifying optimal section combinations for automated novelty prediction in academic papers. Expert Systems with Applications, 273, 126778.

Zaheer, M., Guruganesh, G., Dubey, K. A., Ainslie, J., Alberti, C., Ontanon, S., Pham, P., Ravula, A., Wang, Q., & Yang, L. (2020). Big bird: Transformers for longer sequences. Advances in Neural Information Processing Systems, 33, 17283–17297.

Zhang, C., Tian, L., & Chu, H. (2023). Usage frequency and application variety of research methods in library and information science: Continuous investigation from 1991 to 2021. Information Processing & Management, 60(6), 103507.

Zhang, C., Yan, X., Zhao, L., & Zhang, Y. (2025). Enhancing keyphrase extraction from academic articles using section structure information. Scientometrics, 130(4), 2311–2343.

Zhang, Z., Tam, W., & Cox, A. (2021). Towards automated analysis of research methods in library and information science. Quantitative Science Studies, 2(2), 698–732.

Zhu, M., Weng, Y., Yang, L., & Zhang, Y. (2025). Deepreview: Improving llm-based paper review with human-like deep thinking process. Proceedings of the 63rd Annual Meeting of the Association for Computational Linguistics (Volume 1: Long Papers), 29330–29355.